\documentclass[11pt]{article}

\usepackage[final]{acl}
\usepackage{booktabs}
\usepackage{multirow}
\usepackage{times}
\usepackage{latexsym}
\usepackage{listings}
\usepackage{tcolorbox}
\usepackage{siunitx}
\usepackage[T1]{fontenc}

\usepackage[utf8]{inputenc}

\usepackage{microtype}

\usepackage{inconsolata}

\usepackage{graphicx}
\usepackage{comment}
\usepackage{amsfonts}

\title{Agreement in Representation Space for Open-Ended Self-Consistency}

\author{Paula Ontalvilla \quad Gorka Azkune \quad Aitor Ormazabal \\
       HiTZ Center - Ixa, University of the Basque Country (UPV/EHU)\\
       \texttt{paula.ontalvilla@ehu.eus} }

\begin{document}
\maketitle
\begin{abstract}
Self-consistency improves LLM reasoning by sampling multiple outputs and selecting the most consistent answer, but existing formulations largely rely on exact matching and therefore remain limited to tasks with categorical outputs. In this work, we study self-consistency in open-ended generation tasks such as code synthesis and text summarization. We hypothesize that consistency can be understood as a geometric property of the generation space, where semantically compatible generations concentrate in similar regions of representation space. To study this hypothesis, we introduce Embedding-Based Agreement (EBA), a simple training-free operationalization that estimates agreement by clustering sampled generations in embedding space. Through experiments on mathematical reasoning, code generation, and summarization, we show that agreement in representation space provides a robust and scalable signal of self-consistency for open-ended tasks. In particular, EBA consistently outperforms random selection and exhibits more stable scaling behavior than recent selection approaches based on LLM evaluation or uncertainty estimation. We further show that these agreement signals remain stable across model families and embedding spaces, even with native hidden representations. 
Finally, our analysis shows that the geometric location occupied by sampled generations is strongly correlated with generation quality: generations concentrated near central regions of representation space tend to correspond to more reliable outputs, whereas peripheral generations are substantially less accurate. Overall, our findings support viewing self-consistency as a property of the geometric organization of sampled generations rather than exact symbolic overlap.%
\end{abstract}

\section{Introduction}
\label{sec:intro}
\begin{figure}[htb]
    \centering
    \includegraphics[width=\linewidth]{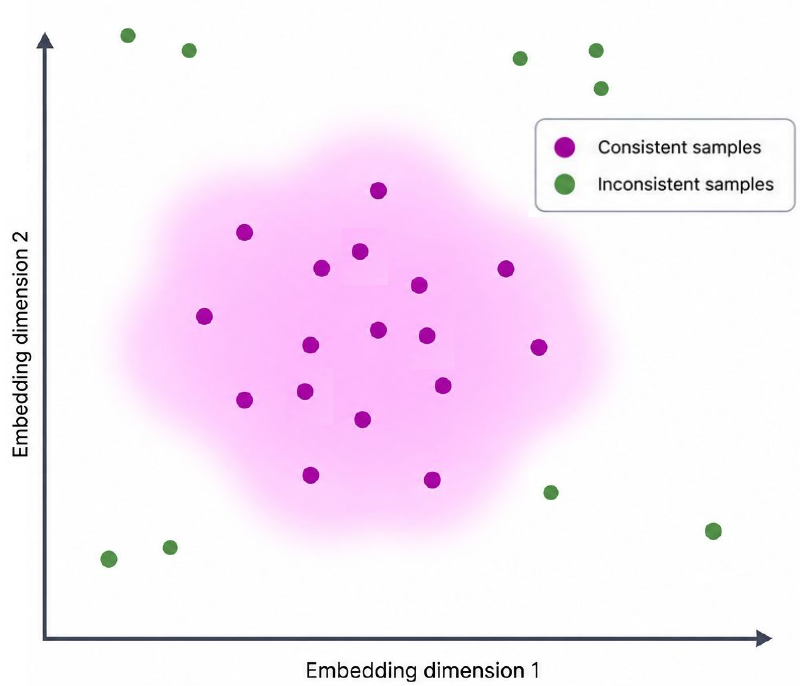}
    \caption{Illustration of our central hypothesis: self-consistency can be understood as a geometric property of the generation space. Semantically consistent generations (purple) are expected to accumulate in dense regions of representation space, forming dominant agreement structures. In contrast, inconsistent generations (green) tend to appear more isolated and sparsely distributed. Under this perspective, agreement can be estimated by identifying dominant concentration regions rather than relying on exact symbolic overlap.}
    \label{fig:example_main}
\end{figure}
Self-consistency has emerged as one of the most effective inference-time strategies for improving the reasoning capabilities of large language models (LLMs). Classical self-consistency methods sample multiple reasoning paths and select the most frequent final answer through majority voting~\citep{wang2022self}. This approach has proven highly effective in tasks with categorical outputs constrained to a limited set of possibilities, such as mathematical reasoning and multiple-choice question answering, where agreement can be naturally defined through exact answer matching.

However, extending self-consistency beyond closed-form tasks remains challenging. In open-ended generation settings such as code generation or summarization, responses that are semantically equivalent may differ substantially at the lexical or structural level, making exact matching ineffective. Recent approaches address this limitation by relying on LLM-based evaluation~\citep{chen2023universal} or uncertainty-based scoring signals~\citep{kang2026scalable}. While effective in some settings, these methods raise broader questions about what agreement actually means in open-ended generation and how consistency signals should be estimated when outputs do not admit canonical symbolic forms.

In this work, we study self-consistency from the perspective of representation-space geometry. We hypothesize that semantically consistent generations occupy similar regions in embedding space and that agreement can therefore be understood as a geometric property of the distribution of sampled generations (see Figure~\ref{fig:example_main}). Building on this view, we introduce Embedding-Based Agreement (EBA)\footnote{The code is publicly available at \url{https://github.com/hitz-zentroa/EBA}}, a simple training-free framework that estimates self-consistency by clustering sampled generations in representation space and selecting representative generations from dominant concentration regions.

Rather than presenting EBA primarily as a new decoding strategy, our goal is to investigate whether agreement in embedding space provides a reliable and scalable signal of self-consistency across both verifiable and open-ended tasks. Through experiments on mathematical reasoning (MATH500), code generation (HumanEval), and summarization (CNN/DM), we show that geometric agreement consistently improves generation selection quality over random sampling and existing selection approaches, and remains robust across model families, model scales, embedding spaces, and native hidden representations.

Our analysis further reveals several properties of agreement in generation space. First, agreement signals strengthen as the number of sampled generations increases, suggesting that consistency emerges from the collective organization of the sampled distribution. Second, embedding-based agreement exhibits more stable scaling behavior than recent approaches based on LLM evaluation or uncertainty estimation. Third, we find that the geometric location occupied by sampled generations strongly correlates with generation quality: generations concentrated near central regions of representation space consistently produce more reliable outputs, whereas peripheral generations are substantially less accurate. Finally, we show that embedding-based agreement closely connects to classical majority-voting self-consistency in closed-form reasoning tasks, suggesting a unified geometric interpretation of self-consistency across categorical and open-ended generation settings.

Overall, our findings suggest that self-consistency can be understood as a property of the geometric organization of sampled generations in representation space, opening a broader perspective on agreement estimation beyond exact symbolic matching.

\section{Related Work}
\label{sec:sota}

Self-consistency (SC)~\citep{wang2022self} improves reasoning in LLMs by sampling multiple reasoning paths and selecting the most frequent final answer through majority voting. This approach has proven highly effective for tasks with categorical and verifiable outputs, such as mathematical reasoning and multiple-choice question answering, where agreement can be determined through exact answer matching.

However, many generation tasks do not naturally admit such limited answer spaces. In open-ended settings such as summarization, code generation, or free-form reasoning, semantically equivalent outputs may differ substantially at the lexical or structural level, making exact-match voting insufficient for estimating consistency.

Several approaches have recently attempted to extend self-consistency beyond closed-form reasoning tasks. Universal Self-Consistency (USC)~\citep{chen2023universal} prompts an LLM to identify the most consistent response among multiple candidates, replacing symbolic voting with model-based evaluation. More recently, lightweight evaluator-free approaches have been proposed. Self-Certainty (SCe)~\citep{kang2026scalable} estimates response quality from token-level uncertainty signals, while similarity-based voting methods~\citep{jiang2025semantic} leverage semantic similarity in embedding space through pairwise voting.%

Our work is closely related to this latter direction, but differs in both motivation and formulation. Prior work primarily treats embeddings as a mechanism for soft answer matching or response ranking. In contrast, we interpret self-consistency itself as a geometric property of the generation space. Specifically, we hypothesize that semantically coherent generations form concentration regions in representation space, enabling agreement to emerge from the collective geometric organization of sampled outputs rather than from pairwise similarity aggregation alone.

More broadly, our work connects to a long line of research showing that neural representation spaces capture meaningful semantic and structural information, enabling applications such as semantic retrieval~\citep{reimers2019sentence, karpukhin2020dense}, clustering~\citep{aggarwal2012survey}, and paraphrase identification~\citep{lan2018neural}. We build on these observations to study agreement in open-ended generation as a property of embedding-space structure rather than discrete symbolic overlap.

\section{Self-Consistency as Agreement in Output Space}
\label{sec:self-consistency-open-ended}

\subsection{Embedding-space agreement}
\label{subsec:embedding-agreement}
Classical self-consistency methods estimate agreement through exact matching or majority voting over categorical outputs. However, in open-ended generation tasks, semantically equivalent responses may differ substantially at the lexical or structural level, making symbolic agreement insufficient for capturing consistency.

In this work, we hypothesize that self-consistency in open-ended generation can instead be understood as a geometric property of the generation space. More specifically, independently sampled generations expressing similar semantic content are expected to occupy nearby regions in a shared representation space, as illustrated in Figure~\ref{fig:example_main}. Under this view, consistency emerges not from exact symbolic overlap, but from the formation of dense semantic regions induced collectively by multiple generations.

Formally, let \(x\) denote an input problem and let $\mathcal{G}(x) = \{g_1, g_2, \dots, g_n\}$ be a set of generations sampled independently from a language model conditioned on \(x\). Let $\phi(g_i) \in \mathbb{R}^d$ denote a vector representation of generation \(g_i\), obtained through an embedding function \(\phi\). The sampled generations therefore induce a geometric structure in representation space: $\Phi(x) = \{\phi(g_1), \phi(g_2), \dots, \phi(g_n)\}$.

Our central hypothesis is that the structure of \(\Phi(x)\) contains information about the latent consistency of the sampled outputs. Semantically coherent generations are expected to form concentrated regions or modes in representation space, while inconsistent or low-quality generations should appear more isolated or fragmented. Under this perspective, self-consistency becomes a problem of identifying dominant semantic regions within the geometry induced by the sampled generations.

This interpretation generalizes the intuition underlying classical self-consistency. In closed-form reasoning tasks, majority voting operates over discrete symbolic answers, implicitly assuming that semantically equivalent outputs collapse to the same representation. Embedding-space agreement extends this idea to continuous and structurally diverse outputs, where semantic similarity must be inferred geometrically rather than symbolically.

\subsection{Operationalizing Agreement in Representation Space}
\label{subsec:operationalization}

Building on this perspective, we operationalize self-consistency in open-ended generation through \textit{Embedding-Based Agreement} (EBA). Given an input prompt $x$, we first sample $N$ candidate generations $\{y_1, \dots, y_N\}$ from a language model and map each generation into a continuous representation space through an embedding function $f(\cdot)$, producing embeddings $\{e_1, \dots, e_N\}$.

Under our hypothesis, agreement among generations is reflected in the geometric organization of this representation space. Rather than requiring exact lexical overlap, semantically compatible generations are expected to accumulate in nearby regions, forming concentration structures that reflect dominant modes of agreement.

To estimate this structure, we cluster the embeddings associated with the sampled outputs. Clusters containing larger numbers of generations are interpreted as regions exhibiting stronger agreement among the sampled responses. In this sense, clustering plays a role analogous to majority voting in classical self-consistency: instead of counting identical symbolic answers, agreement is estimated through the accumulation of semantically similar generations in representation space. We then identify the dominant cluster and select as final prediction the sample located closest to its centroid, favoring generations that are most representative of the dominant agreement region.

In our implementation, clustering is performed using agglomerative clustering~\citep{mullner2011modern}, with the number of clusters determined automatically through silhouette analysis~\citep{rousseeuw1987silhouettes}. Unless otherwise specified, embeddings are obtained using pretrained embedding models independently from the generation model itself (implementation details are provided in Appendix~\ref{app:implementation}).

\section{Experimental Setup}
\label{sec:setup}

\begin{figure*}[htb]
    \centering
    \includegraphics[width=\linewidth]{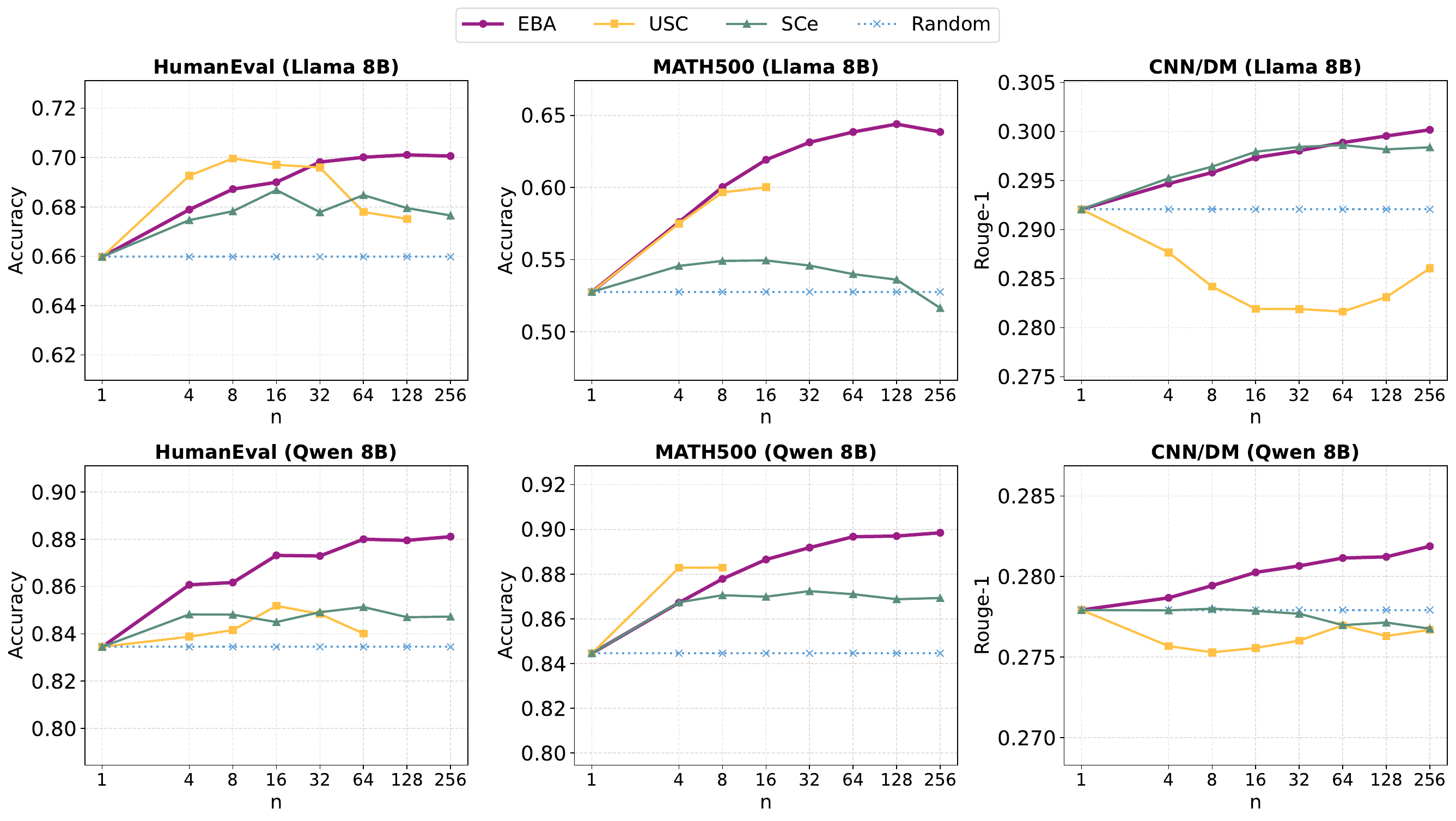}
    \caption{Performance of EBA, random selection, Universal Self-Consistency (USC), and self-certainty (SCe) across increasing numbers of sampled generations on HumanEval, MATH500 and CNN/DM using Llama 3-8B (top) and Qwen 3-8B (bottom). EBA consistently improves with increasing numbers of samples and tends to plateau at higher values, whereas USC and SCe generally plateau earlier or even degrade. This suggests more stable scaling behaviour for embedding-based agreement.}
    \label{fig:main}
\end{figure*}

\paragraph{Datasets.}
We evaluate Embedding-Based Agreement (EBA) on three generation tasks covering both verifiable and open-ended settings: HumanEval~\citep{chen2021evaluating} for code generation, CNN/DailyMail (CNN/DM)~\citep{cnn_dm1, cnn_dm2} for summarization,  and MATH500~\citep{hendrycks2021measuring,lightman2023lets} for mathematical reasoning. We report accuracy for HumanEval and MATH500, and Rouge-1~\citep{lin2004rouge} for CNN/DM. Unless otherwise specified, agreement on MATH500 is computed directly over full generations.%

\paragraph{Models.}
Our main experiments use Llama~3.1-8B~\citep{grattafiori2024llama} and Qwen~3-8B~\citep{qwen3technicalreport} as generation models, together with Qwen-Embedding-8B~\citep{zhang2025qwen3} as embedding function \(\phi\) for representation extraction.

\paragraph{Reference methods.}
We compare EBA against Universal Self-Consistency (USC)~\citep{chen2023universal}, Self- Certainty (SCe)~\citep{kang2026scalable}, and random selection.

\paragraph{Implementation details.}
Agreement regions are estimated through agglomerative clustering with cosine distance, where the number of clusters is selected automatically using the Silhouette score. We use a sampling temperature of $T=0.7$ with $top\_p=1.0$ and report results for $n$ between $1$ and $256$.\footnote{For USC, the evaluator model can run out of context for large $n$ and tasks with long generations, in which case we report results up to the context limit.} To reduce noise in the results, we use 50 bootstrapping iterations for every reported result, drawing sets of generations from a fixed pool of 512 sampled generations for each input. For USC, we use as the selector the same LLM that was used to generate the samples. Additional implementation details, prompts, and hardware specifications are provided in Appendix~\ref{app:implementation} and Appendix~\ref{app:prompts}.

\section{Empirical Evidence of Embedding-Space Agreement}
\label{sec:evidence}
This section evaluates agreement in representation space as a signal of self-consistency in open-ended generation. We study the effectiveness, scaling behavior, and robustness of Embedding-Based Agreement (EBA) across models and representation spaces (full results in Appendix \ref{app:full_results}).

\subsection{Agreement in embedding space improves selection quality}
\label{subsec:selection}

Figure~\ref{fig:main} shows the performance of Embedding-Based Agreement (EBA) across code generation (HumanEval), mathematical reasoning (MATH500), and summarization (CNN/DM), together with several reference selection methods. Across both Llama and Qwen models, EBA consistently outperforms random selection, indicating that agreement in representation space provides a substantially stronger signal than selecting arbitrary generations from the sampled set.

The improvements are particularly pronounced in code generation and mathematical reasoning, where EBA improves over random selection by up to $4$ and $17$ points respectively, depending on the model. The effect is smaller but still consistent for summarization (approximately $+1$ Rouge-1 point), where generations tend to occupy more concentrated regions of the output space and evaluation metrics may under-reward semantically equivalent paraphrastic outputs. Overall, these results suggest that agreement can emerge from the geometric organization of sampled generations even when responses differ lexically or structurally, supporting the view of self-consistency as a property of the generation distribution rather than exact symbolic matching.

\subsection{Agreement strengthens with increased sampling}
\label{subsec:scaling}

As shown in Figure~\ref{fig:main}, EBA exhibits consistent and near-monotonic improvements as the number of sampled generations increases across datasets and model families. The effect is particularly pronounced in HumanEval and MATH500, where performance tends to improve over a broad range of sampling budgets before gradually leveling off at larger values of $n$. In contrast, summarization exhibits smaller but still positive gains as the number of generations increases.

These results suggest that agreement signals become increasingly informative as the generation distribution is explored more thoroughly. As additional generations are sampled, dominant concentration regions in representation space may become more clearly defined, making agreement estimates more reliable. The observed scaling behavior supports the view of self-consistency as a geometric property emerging from the collective organization of sampled generations rather than exact lexical overlap between outputs.%

\subsection{Geometric agreement provides a robust signal of self-consistency}
\label{subsec:usc}

Figure~\ref{fig:main} additionally compares EBA against two recent approaches for extending self-consistency beyond exact matching: Universal Self-Consistency (USC) and Self-Certainty (SCe). While both methods generally improve over random selection, EBA exhibits more stable behavior across tasks and sampling regimes.

USC frequently degrades as the number of candidate generations increases, for example it loses around two points on HumanEval for the Llama model. This behavior suggests increasing difficulty in reliably identifying agreement through direct LLM-based comparison over larger candidate sets. Moreover, USC is inherently constrained by the maximum context length of the underlying model, limiting its scalability to large numbers of generations.

SCe demonstrates substantially stronger scaling behavior than USC and remains competitive across several settings. However, its performance is less stable as sampling increases; for example, on MATH500 with Llama~8B, performance peaks early before dropping by roughly four points between $n=16$ and $n=256$. In contrast, EBA consistently matches or outperforms SCe on MATH500 while remaining competitive on open-ended tasks such as HumanEval and CNN/DM. Overall, these results suggest that geometric agreement provides a robust and scalable signal of self-consistency across diverse generation settings.

\subsection{Agreement is stable across model scales and embedding spaces}
\label{subsec:agreement-stable}

\begin{figure*}[htb]
    \centering
    \includegraphics[width=\linewidth]{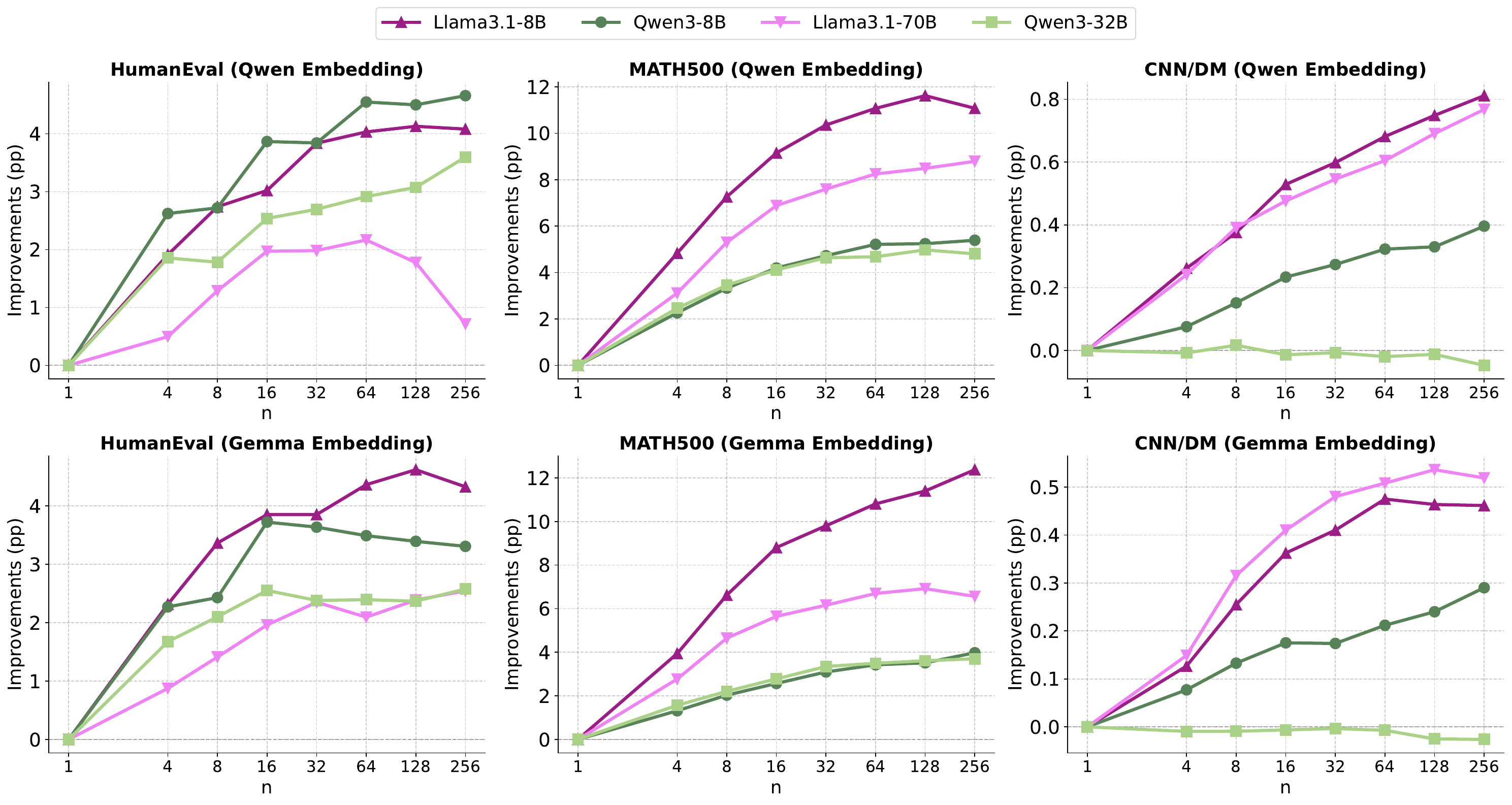}
    \caption{Improvement of embedding-based agreement (EBA) as the number of sampled generations increases across code generation (HumanEval), summarization (CNN/DM), and mathematical reasoning (MATH500). The y-axis reports gain in \textbf{percentage points (pp)} over random performance. Results are shown for multiple language models and embedding representations, including Qwen and Gemma embeddings. Across model scales and embedding spaces, EBA exhibits consistent scaling behavior, indicating that agreement in embedding space reflects a robust and model-agnostic property of the generation distribution.}
    \label{fig:5_4_scale_and_embeddings}
\end{figure*}

Figure~\ref{fig:5_4_scale_and_embeddings} analyzes the robustness of EBA across different language model scales and embedding spaces. We evaluate both larger generation models (Llama 70B and Qwen 32B) and alternative embedding models (Qwen and Gemma embeddings~\citep{vera2025embeddinggemma}). Across all configurations, the same qualitative behavior observed in Figure~\ref{fig:main} remains consistent: agreement-based selection continues to outperform random selection and benefits from increased sampling.

While absolute performance and improvement magnitudes vary across models and embedding spaces, several qualitative patterns remain consistent. In particular, EBA exhibits strong and largely monotonic gains on MATH500 and generally positive trends on CNN/DM, whereas HumanEval displays more variable behavior, particularly for larger models. Larger models also tend to exhibit smaller improvements overall, which may be explained by their stronger baseline performance and consequently smaller potential gains from selection. These results indicate that the proposed framework captures a general property of the geometry of sampled generations rather than an artifact of a particular representation space.

\begin{figure*}[htb]
    \centering
    \includegraphics[width=\linewidth]{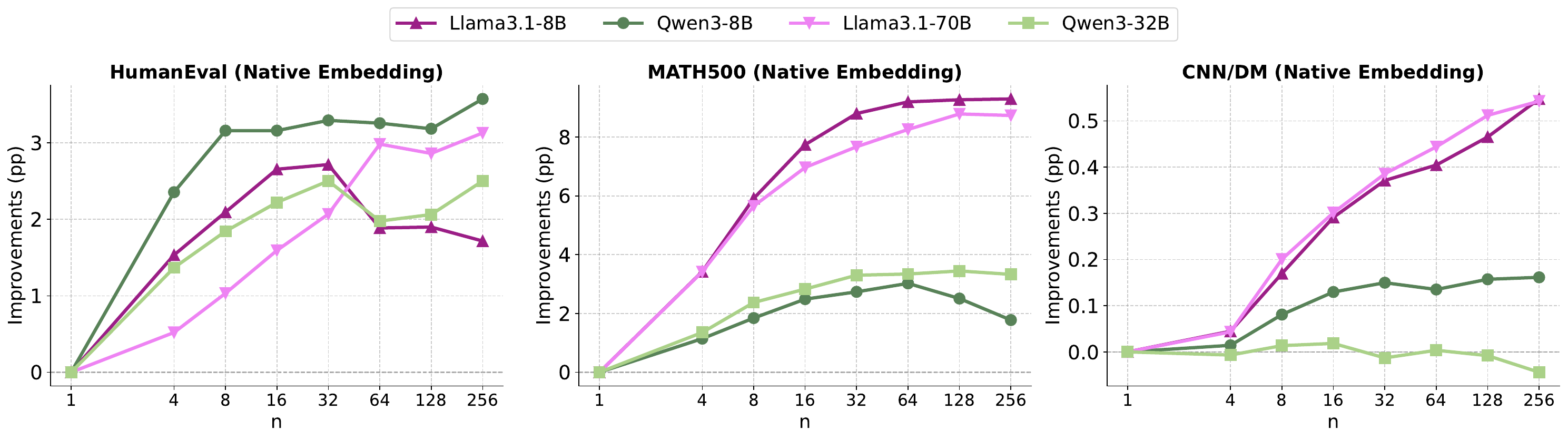}
    \caption{Improvement of embedding-based agreement (EBA) using native hidden representations as the number of sampled generations increases across all datasets.  The y-axis reports gain in \textbf{percentage points (pp)} over random performance. Results are shown for multiple language models. }
    \label{fig:5_5_native_embeddings}
\end{figure*}

\subsection{Agreement emerges in native hidden representations}
\label{subsec:native-embs}

Figure~\ref{fig:5_5_native_embeddings} evaluates EBA using representations extracted directly from the generation model itself, replacing external embedding models with the final sampled token's hidden representation. Across all datasets and model families, EBA continues to exhibit the same qualitative behavior observed with pretrained embedding models: agreement-based selection consistently improves with increased sampling and remains substantially stronger than random selection.

Although performance is typically slightly lower than with specialized embedding models, with differences often within ±1–2 points across datasets, several qualitative patterns remain similar. Improvements are generally stronger and more stable for MATH500 and CNN/DM, while HumanEval exhibits more variable behavior, particularly at larger sampling budgets. These results suggest that useful agreement signals may already be present in native hidden representations and do not rely exclusively on external semantic embedding spaces. More broadly, they are consistent with the view that aspects of self-consistency may be reflected in the internal organization of sampled generations in representation space.

\section{Analysis of Self-Consistency in Open-Ended Generation}
\label{sec:analysis}
Results from Section~\ref{sec:evidence} show that agreement in representation space is a strong signal of self-consistency across tasks. We further study the underlying geometry and its relationship to self-consistency, as well as the effects of local and global structure on generation quality (full results in Appendix~\ref{app:full_results}).

\subsection{Connection to closed-form self-consistency}
\label{subsec:connection-closed-form}

Classical self-consistency operates in discrete answer spaces, where sampled generations are reduced to final answers and aggregated through majority voting. In contrast, EBA targets open-ended generation settings in which outputs cannot be naturally mapped to a small set of categorical responses. A natural question, therefore, is whether embedding-based agreement remains connected to the original intuition underlying closed-form self-consistency.

To study this connection, we evaluate EBA on MATH500 under two representation settings: (i) using the complete generation, including both reasoning trace and final answer, and (ii) using only the extracted final answer. We compare both variants against standard majority-voting self-consistency.

\begin{figure}
    \centering
    \includegraphics[width=\linewidth]{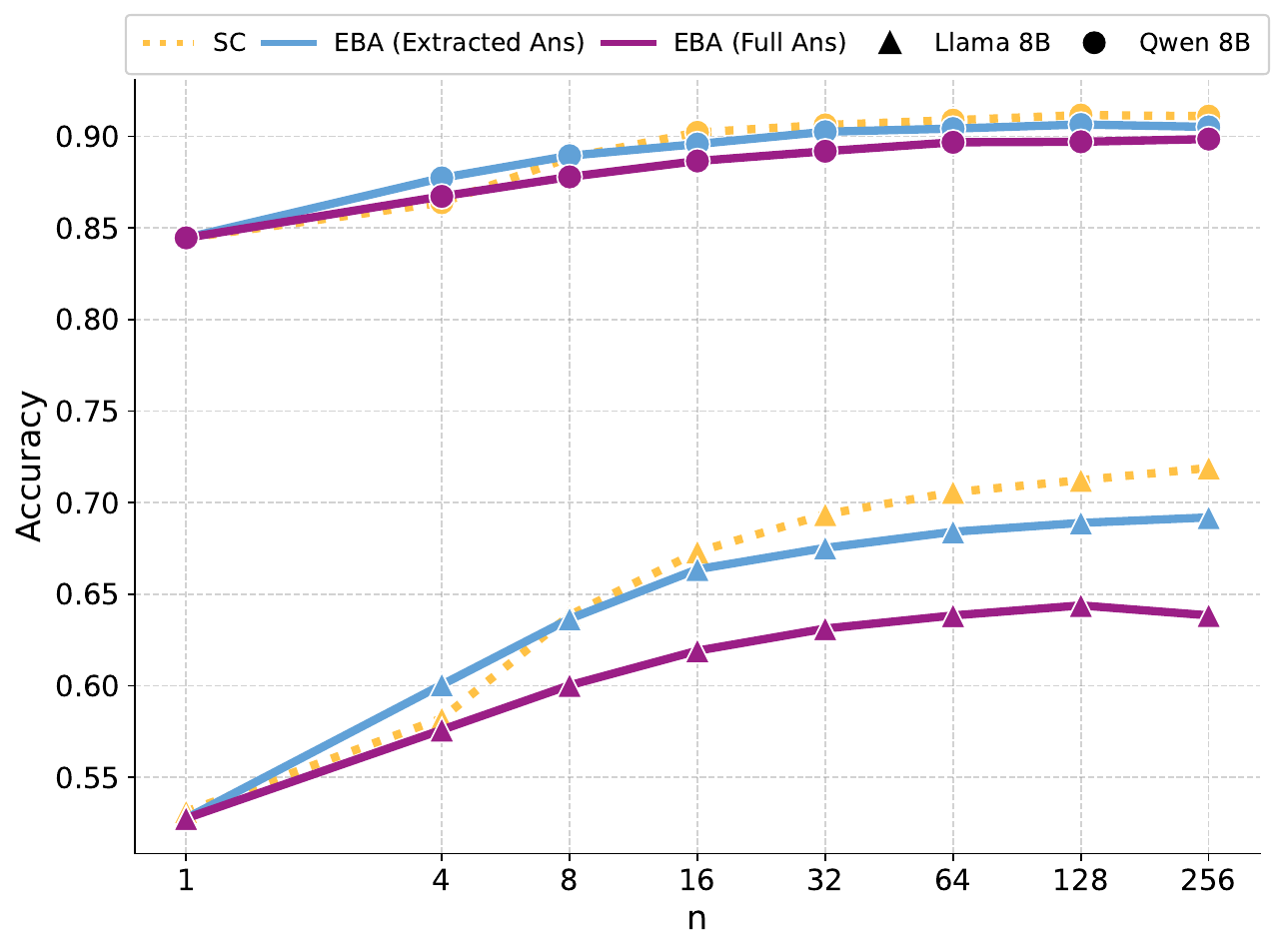}
    \caption{Comparison of self-consistency (SC) and embedding-based aggregation method using the complete generation (EBA (Full Ans)) and using only the extracted final answer (EBA (Extracted Ans)) on MATH500. Results are shown for Llama 8B and Qwen 8B models.}
    \label{fig:closed-form-sc}
\end{figure}

Figure~\ref{fig:closed-form-sc} shows that embedding-based agreement exhibits scaling behavior remarkably similar to classical self-consistency. Across both Llama 3-8B and Qwen 3-8B, performance improves consistently as additional generations are sampled, closely following the monotonic trend of majority voting. This similarity is particularly strong when agreement is computed over isolated final answers, where EBA approaches the performance of exact-match self-consistency across all sample sizes. Similar trends are observed for larger models (see Appendix~\ref{app_math_bigger_models}).

The comparison between full-generation and answer-only agreement further highlights an important property of open-ended generation spaces. While both variants exhibit similar scaling dynamics, agreement over complete generations consistently underperforms answer-level agreement, suggesting that embedding representations remain sensitive to differences in reasoning structure, lexical choice, and chain-of-thought organization even when generations lead to the same final answer.

Overall, these findings suggest that EBA can be viewed as a continuous extension of classical self-consistency. When outputs become effectively discrete, agreement in representation space behaves similarly to majority voting, while naturally extending the same idea to more diverse generation settings.

\subsection{Generation quality correlates with geometric structure}
\label{subsec:global-centroid}
\begin{figure*}[htb]
    \centering
    \includegraphics[width=0.9\linewidth, height=0.36\linewidth]{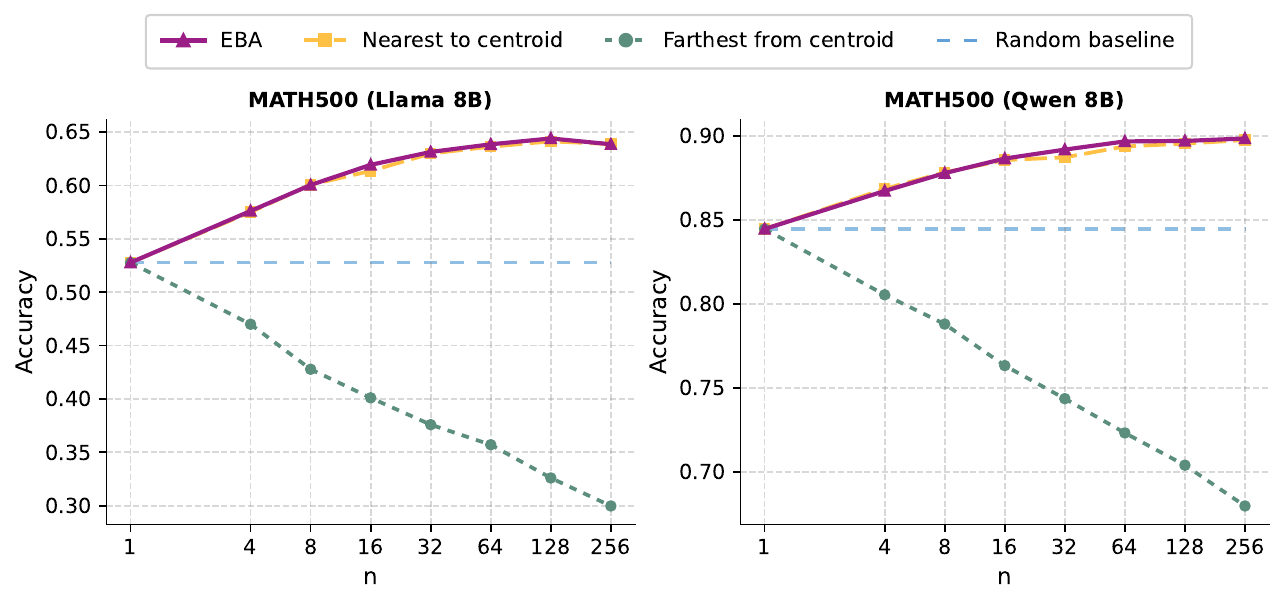}
    \caption{Comparison of generation selection strategies on MATH500 for Llama 8B and Qwen 8B as the number of sampled generations $(n)$ increases. Selecting generations closest to the global centroid achieves performance nearly similar to EBA, whereas selecting generations farthest from the centroid causes substantial performance degradation. The dashed line denotes the random baseline corresponding to performance at $n=1$.}
    \label{fig:global_centroid}
\end{figure*}
While EBA operationalizes agreement through clustering, this discretization is not the only possible way to estimate agreement in continuous representation spaces. An alternative view is to treat agreement as a global geometric property of the sampled generations without explicitly partitioning the space into clusters. Under this perspective, generations located near the global centroid may correspond to semantically central or highly representative outputs, whereas generations located farther away may reflect more isolated or inconsistent behaviors. Figure~\ref{fig:global_centroid} compares EBA against two centroid-based selection strategies on MATH500: selecting the generation closest to the global centroid and selecting the generation farthest from it.

The results reveal a striking geometric pattern. Across both Llama and Qwen models, selecting generations closest to the global centroid achieves performance remarkably similar to EBA across all sampling regimes. In contrast, selecting generations farthest from the centroid produces dramatic degradations in accuracy, reaching drops of nearly 20 points for large sample sizes. These results indicate that the geometric location occupied by a generation in representation space is strongly correlated with its quality: generations concentrated near central regions tend to correspond to higher-quality outputs, whereas peripheral generations are substantially less reliable.

At the same time, the near-equivalence between EBA and global-centroid selection suggests that the explicit discretization introduced by clustering contributes only marginally under the current experimental setting. Several factors may explain this behavior. Prior work~\citep{liang2022mind} has shown that embedding representations often populate narrow high-dimensional conic regions, which may reduce the separability of distinct semantic modes. Additionally, the relatively simple clustering strategy used in this work may fail to capture finer-grained structure in the generation distribution. More broadly, open-ended generation spaces may not exhibit sharply separated semantic regions, but rather smoother concentration gradients organized around dominant semantic directions.

\section{Conclusion}
\label{sec:conclusion}
In this work, we studied self-consistency beyond categorical answer spaces. We hypothesized that semantically compatible generations concentrate in nearby regions of embedding space, and introduced Embedding-Based Agreement (EBA), a simple training-free framework that estimates agreement by clustering sampled generations and selecting representative outputs from dominant concentration regions.

Across mathematical reasoning, code generation, and summarization, we showed that agreement in embedding space provides a reliable and scalable selection signal. EBA consistently improves over random selection and exhibits more stable scaling behavior than recent approaches based on LLM evaluation or uncertainty estimation. Moreover, the observed agreement patterns remain robust across model families, model scales, embedding spaces, and native hidden representations.

We further connected EBA to classical self-consistency on closed-form reasoning tasks, showing that EBA closely follows the scaling behavior of exact-match majority voting while naturally extending to open-ended generation. Our analysis also reveals that the geometric location occupied by sampled generations is strongly correlated with generation quality, with central regions in representation space consistently associated with more reliable outputs than peripheral regions.

Overall, our findings support viewing self-consistency as a geometric property of sampled generations rather than a phenomenon tied exclusively to exact symbolic overlap. This perspective provides a unified interpretation of agreement across discrete and open-ended tasks, and opens new directions for inference-time generation selection based on representation-space structure.

\section*{Limitations}
\label{sec:limitations}

While our results suggest that embedding-based agreement captures a meaningful signal of self-consistency in open-ended generation, several limitations remain.

\paragraph{Representation geometry and semantic alignment.}
Our approach assumes that semantically compatible generations occupy nearby regions in representation space. Although this assumption holds sufficiently well to produce consistent improvements across tasks, the results of Section~6 suggest that current embedding spaces exhibit relatively low-separation geometric structure, with generations often concentrating in a small number of broad regions. This may reflect known properties of neural embedding spaces, such as anisotropy and concentration in narrow high-dimensional cones. As a result, embedding proximity does not always perfectly correspond to semantic equivalence, potentially limiting the precision of agreement estimation.

\paragraph{Limitations of clustering-based agreement estimation.}
EBA operationalizes agreement through agglomerative clustering and heuristic model-selection criteria based on silhouette analysis. While this provides a simple and fully training-free mechanism for estimating dominant concentration regions, the relatively small gap observed between EBA and global centroid selection suggests that current clustering strategies may not fully capture the local structure of the generation space. More expressive geometric modeling or representation-learning approaches may therefore yield improved estimates of semantic agreement.

\paragraph{Computational considerations.}
Although EBA avoids additional LLM evaluation calls and remains substantially cheaper than evaluator-based approaches such as USC, it still requires generating multiple samples and computing their representations. For large sampling budgets or long generations, embedding extraction and storage may introduce additional computational overhead. At the same time, our experiments suggest that simpler alternatives based on native embeddings may provide attractive practical trade-offs.

\paragraph{Scope of evaluation.}
Our experiments focus on summarization, code generation, and mathematical reasoning as representative open-ended generation settings. While these tasks cover both verifiable and non-verifiable outputs, additional evaluation on domains such as dialogue, long-form generation, or agentic reasoning would be necessary to fully assess the generality of the proposed perspective.

\bibliography{custom}

\appendix

\section{Implementation details}\label{app:implementation}

\paragraph{Clustering.} We use the same temperature and sampling strategy for all model sizes. Specifically, we generate  512 samples per input using temperature $T=0.7$ and $top\_p=1.0$ sampling in a zero-shot setting.  We then create subsets of size 4, 8, 16, 32, 64, 128, and 256. We do not report experiments for $n=512$ because we perform 50 bootstrap resamples, which would not be feasible when using all 512 samples. We use the vLLM framework for generation, whereas native embeddings are extracted using Hugging Face, as it provides access to hidden-dimensional representations.

For all datasets except the HumanEval dataset, clustering is applied directly to the generated responses, which are also the embedded representations. For HumanEval, clustering is instead performed on the extracted final answers, which are likewise used as the embedded representations. Regarding the embedding model, we use the generations produced by each model independently as input. Hierarchical clustering~\citep{mullner2011modern}\footnote{\url{https://scikit-learn.org/stable/modules/generated/sklearn.cluster.AgglomerativeClustering.html}} is performed by constructing the complete clustering tree and selecting the partition that maximizes the Silhouette score. We use average linkage and cosine distance, as embeddings are computed using cosine similarity. All the code is available at the projects' repository\footnote{\url{https://github.com/hitz-zentroa/EBA}}.

\paragraph{Reference methods}
For USC, we use the temperature values recommended for each model to obtain the best performance, i.e. for Llama models ($T=0.7$, $top\_p=0.8$, $top\_k=20$) and for Qwen models ($T=0.6$, $top\_p=0.9$).  We use the prompts from the original paper~\cite{chen2023universal} with minor modifications depending on the dataset (see Appendix~\ref{app:prompts} ). When USC fails to choose one answer we randomly select one. 

For SCe, we use the authors’ released implementation\footnote{\url{https://github.com/backprop07/Self-Certainty}} while adapting it to our experimental setup, i.e., retaining their self-certainty scoring and selection procedure while applying it to our generated samples. We also use a bootstrapping of 50 for these experiments.

\paragraph{Datasets.}
All datasets are publicly available\footnote{\url{https://huggingface.co/datasets/HuggingFaceH4/MATH-500}}\footnote{\url{https://huggingface.co/datasets/openai/openai_humaneval}}\footnote{\url{https://huggingface.co/datasets/abisee/cnn_dailymail}}. Due to  computational constraints, we use a subset of 1,000 samples for CNN/DM from the 3.0.0 version. Specifically, we select the first 1,000 samples of the dataset.

For evaluation in the case of MATH dataset, we adopted an LLM-based evaluation, in line with a widely-used library\footnote{\url{https://github.com/openai/simple-evals}}. Since both predictions and gold answers often contain \LaTeX{} expressions and certain expressions can be written in different but mathematically equivalent ways (e.g., \lstinline|\frac{2}{3}| and \lstinline|2/3|), string matching is not always a reliable metric. To address this, we used the Qwen 8B model for evaluation. This choice was supported by an empirical study in which we measured the agreement between Qwen 8B and human-annotated labels on 200 instances, achieving a 95.5\% match. For HumanEval dataset, we first, extract the final answer with a combination of filters and regular expressions and then, we evaluate the function with the official evaluation harness\footnote{\url{https://github.com/openai/human-eval}} which runs the function against a set of unit tests.

\paragraph{Computational Infrastructure.}
All experiments were conducted using NVIDIA A100 GPUs. We use 1 GPU for small model inference (Llama 8B and Qwen 8B), 2 GPUS for Qwen 32B and 4 GPUs for Llama70B model inference. For both embeddings models we use 1 GPU and clustering was made in CPU.

\paragraph{Licences.} To the best of our knowledge, all the data sources we use are available for non-commercial use.

\begin{itemize}
  \item \textbf{MATH}~\citep{hendrycks2021measuring, lightman2023lets}: It is under a MIT License\footnote{\url{https://github.com/openai/prm800k/tree/main?tab=MIT-1-ov-file}}.
  \item \textbf{CNN/DailyMail}~\citep{cnn_dm1, cnn_dm2}: It is under Apache-2.0 license\footnote{\url{https://huggingface.co/datasets/abisee/cnn_dailymail\#licensing-information}}.
  \item \textbf{HumanEval}~\citep{chen2021evaluating}: It is under a MIT License\footnote{\url{https://github.com/openai/human-eval?tab=MIT-1-ov-file}}.
\end{itemize}

Regarding the models we used:
\begin{itemize}
  \item \textbf{Llama models}~\citep{grattafiori2024llama}: They are licensed under the Llama 3.1 Community License, Copyright © Meta Platforms, Inc. All Rights Reserved\footnote{\url{https://huggingface.co/meta-llama/Llama-3.1-70B-Instruct/blob/main/LICENSE}}.
  \item \textbf{Qwen models}~\citep{qwen3technicalreport, zhang2025qwen3}: They are publicly accessible under Apache 2.0\footnote{\url{https://github.com/QwenLM/Qwen3\#license-agreement}}\footnote{\url{https://huggingface.co/Qwen/Qwen3-Embedding-8B/blob/main/LICENSE}}.
  \item \textbf{Gemma Embeddings model}~\citep{vera2025embeddinggemma}: It is under Gemma License\footnote{\url{https://ai.google.dev/gemma/terms}}.
\end{itemize}

\section{Prompts}\label{app:prompts}
We use the same prompts for all models, while using different prompts for dataset.  For USC  We use the prompts from the original paper~\cite{chen2023universal} with minor modifications depending on the dataset.

\section{Analysis of Self-Consistency in Open-Ended Generation for Larger Models}\label{app_math_bigger_models}

\begin{figure}[htb]
    \centering
    \includegraphics[width=\linewidth]{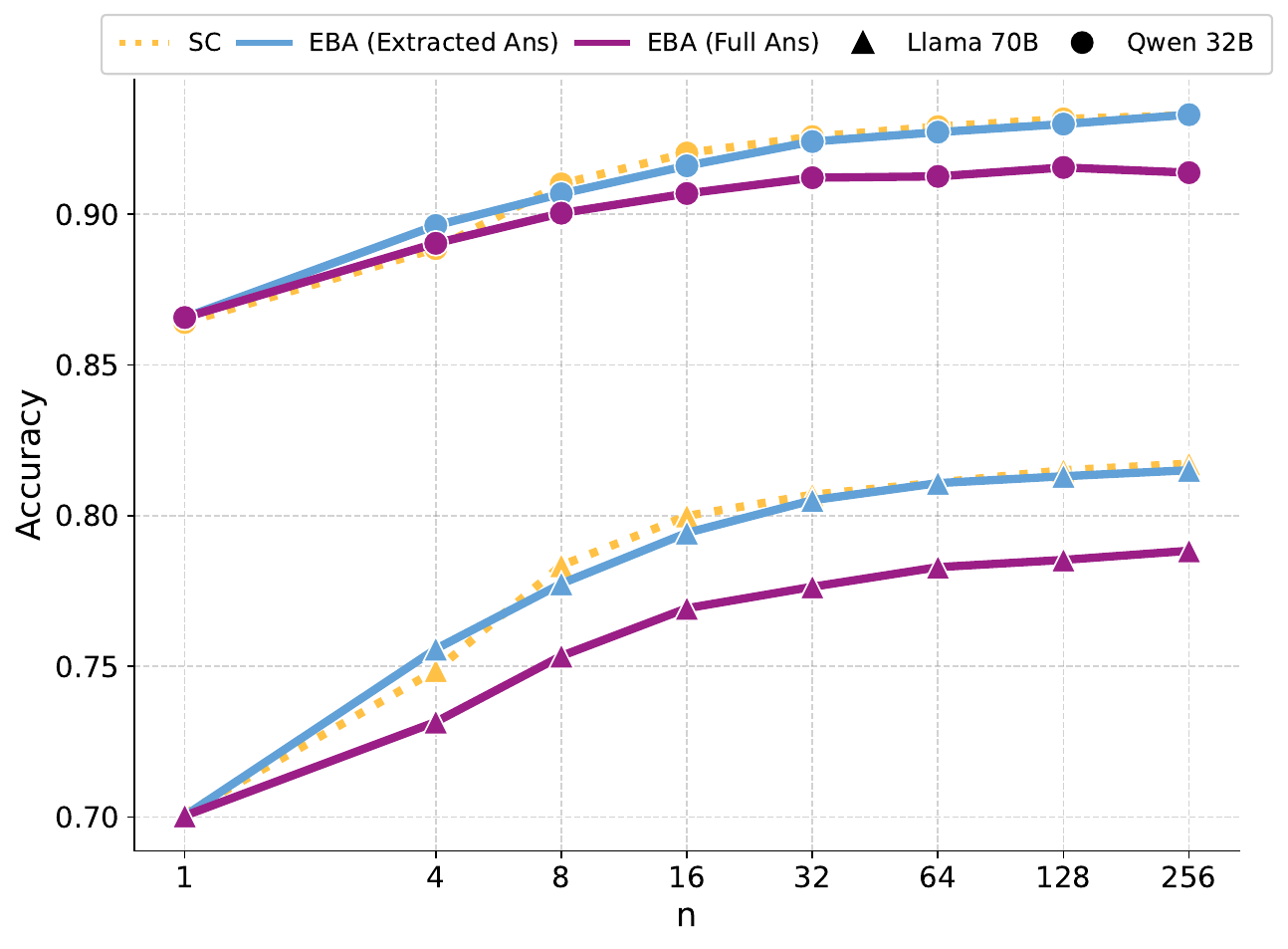}
    \caption{Comparison of self-consistency (SC) and embedding-based aggregation method using the complete generation (EBA (Full Ans)) and using only the extracted final answer (EBA (Extracted Ans)) on MATH500. Results are shown for Llama 70B and Qwen 32B models.}
    \label{fig:closed-form-sc-large-scale}
\end{figure}

We replicate the MATH500 evaluation using larger backbone models to test the scalability of embedding-based agreement (EBA). Results show (see Figure~\ref{fig:closed-form-sc-large-scale}) the same qualitative behavior: EBA improves monotonically with additional samples and closely follows classical self-consistency trends across all model sizes.

As in smaller models, agreement computed over final answers is more closely aligned with majority voting than full-generation embeddings, although both exhibit consistent scaling patterns. These findings confirm that the relationship between embedding-space agreement and self-consistency remains stable as model capacity increases.

\begin{figure*}[htb]
    \centering
    \begin{tcolorbox}[colback=gray!5, colframe=black!60, boxrule=0.5pt, arc=2mm, left=2mm, right=2mm, title={Prompt for CNN/DailyMail}]
    Summarize the following text in less than 4 sentences. Generate a concise and coherent summary that captures the main points of the text:\\
    \texttt{<INPUT>}\\
    \\
    Summary:\\
    \end{tcolorbox}
\end{figure*}

\begin{figure*}[htb]
    \centering
    \begin{tcolorbox}[colback=gray!5, colframe=black!60, boxrule=0.5pt, arc=2mm, left=2mm, right=2mm, title={Prompt for USC for CNN/DailyMail}]   
    I have the following text: \texttt{<INPUT>} \\
    I have generated the following summaries of the text:
     \texttt{<RESPONSES>}\\
    
    Evaluate the responses.\\
    Select the most consistent response based on majority consensus.\\
    Start your answer with "The most consistent response is Response X" (without quotes).\\
    \end{tcolorbox}
\end{figure*}

\begin{figure*}[htb]
    \centering
    \begin{tcolorbox}[colback=gray!5, colframe=black!60, boxrule=0.5pt, arc=2mm, left=2mm, right=2mm, title={Prompt for HumanEval}]
    \texttt{<INPUT>}\\
    \end{tcolorbox}
\end{figure*}

\begin{figure*}[htb]
    \centering
    \begin{tcolorbox}[colback=gray!5, colframe=black!60, boxrule=0.5pt, arc=2mm, left=2mm, right=2mm, title={Prompt for USC for HumanEval}]   
   I have generated the following code-completion outputs for this function:      \\
   \texttt{<INPUT>} \\
     \texttt{<RESPONSES>}\\
    
    Evaluate the responses.\\
    Select the most consistent response based on majority consensus.\\
    Start your answer with "The most consistent response is Response X" (without quotes).\\
    \end{tcolorbox}
\end{figure*}

\begin{figure*}[htb]
    \centering
    \begin{tcolorbox}[colback=gray!5, colframe=black!60, boxrule=0.5pt, arc=2mm, left=2mm, right=2mm, title={Prompt for MATH500}]
    Solve the following math problem step by step. The last line of your response should be of the form Answer: \$ANSWER (without quotes) where \$ANSWER is the answer to the problem.\\
    \\
    \texttt{<INPUT>}\\
    \\
    Remember to put your answer on its own line after "Answer:", and you do not need to use a \textbackslash \textbackslash boxed command.\\
    \end{tcolorbox}
\end{figure*}

\begin{figure*}[htb]
    \centering
    \begin{tcolorbox}[colback=gray!5, colframe=black!60, boxrule=0.5pt, arc=2mm, left=2mm, right=2mm, title={Prompt for USC for MATH500}]   
  I have generated the following responses to the mathematical problem:     \\
   \texttt{<INPUT>} \\
     \texttt{<RESPONSES>}\\
    
    Evaluate the responses.\\
    Select the most consistent response based on majority consensus.\\
    Start your answer with "The most consistent response is Response X" (without quotes).\\
    \end{tcolorbox}
\end{figure*}

\section{Full results}\label{app:full_results}
This appendix includes full tabular versions of the figures. Tables~\ref{tab:humaneval}, \ref{tab:math500}, and \ref{tab:cnn} present the results of Figure~\ref{fig:main}. Tables~\ref{tab:eba_humaneval}, \ref{tab:eba_math500}, and \ref{tab:eba_cnn_dm} present the results of Figures~\ref{fig:5_4_scale_and_embeddings} and \ref{fig:5_5_native_embeddings}. Table~\ref{tab:matH_sc_8b} and Table~\ref{tab:matH_sc_70b} present the results for Figure~\ref{fig:closed-form-sc} and Figure~\ref{fig:closed-form-sc-large-scale} respectively. Finally, Table~\ref{tab:matH_centroid} presents the result of the Figure~\ref{fig:global_centroid}.

\begin{table*}[ht]
  \centering
  \small
   \resizebox{\textwidth}{!}{
   
  \begin{tabular}{llcccccccc}
    \toprule
    \textbf{Model} & \textbf{Method} & $n=1$ & $n=4$ & $n=8$ & $n=16$ & $n=32$ & $n=64$ & $n=128$ & $n=256$ \\
    \midrule
    \textbf{Llama 8B} & EBA & 65.98 $\pm$ \scriptsize{2.38} & 67.89 $\pm$ \scriptsize{2.23} & 68.72 $\pm$ \scriptsize{2.12} & 69.00 $\pm$ \scriptsize{1.97} & 69.82 $\pm$ \scriptsize{1.54} & 70.01 $\pm$ \scriptsize{1.96} & 70.11 $\pm$ \scriptsize{1.52} & 70.06 $\pm$ \scriptsize{1.58} \\
     & SCe & 65.98 $\pm$ \scriptsize{2.38} & 67.46 $\pm$ \scriptsize{0.78} & 67.83 $\pm$ \scriptsize{0.47} & 68.68 $\pm$ \scriptsize{0.57} & 67.79 $\pm$ \scriptsize{0.43} & 68.48 $\pm$ \scriptsize{0.47} & 67.95 $\pm$ \scriptsize{0.36} & 67.65 $\pm$ \scriptsize{0.51} \\
     & USC & 65.98 $\pm$ \scriptsize{2.38} & 69.27$\pm$ \scriptsize{2.05}& 69.96$\pm$\scriptsize{1.99} & 69.71$\pm$\scriptsize{2.02} & 69.60 $\pm$\scriptsize{1.79}& 67.79$\pm$ \scriptsize{2.17}& 67.51 $\pm$\scriptsize{2.10}& -- \\
     & Random &65.98  & 65.98 &65.98 & 65.98  &65.98  & 65.98 &65.98  & 65.98 \\
    \midrule
    \textbf{Qwen 8B} & EBA & 83.45 $\pm$ \scriptsize{2.10} & 86.07 $\pm$ \scriptsize{1.85} & 86.17 $\pm$ \scriptsize{1.46} & 87.32 $\pm$ \scriptsize{1.17} & 87.29 $\pm$ \scriptsize{1.26} & 88.00 $\pm$ \scriptsize{1.38} & 87.95 $\pm$ \scriptsize{1.21} & 88.11 $\pm$ \scriptsize{1.31} \\
     & SCe & 83.45 $\pm$ \scriptsize{2.10} & 84.82 $\pm$ \scriptsize{0.37} & 84.81 $\pm$ \scriptsize{0.32} & 84.50 $\pm$ \scriptsize{0.27} & 84.91 $\pm$ \scriptsize{0.22} & 85.13 $\pm$ \scriptsize{0.40} & 84.70 $\pm$ \scriptsize{0.31} & 84.73 $\pm$ \scriptsize{0.18} \\
     & USC & 83.45 $\pm$ \scriptsize{2.10} & 83.88 $\pm$ \scriptsize{1.87} & 84.16 $\pm$ \scriptsize{1.75} & 85.18 $\pm$ \scriptsize{1.59} & 84.84 $\pm$ \scriptsize{1.90} & 84.01 $\pm$ \scriptsize{2.13} & -- & -- \\
     & Random & 83.45 & 83.45 & 83.45 & 83.45 & 83.45 & 83.45 & 83.45 & 83.45 \\
    \bottomrule
  \end{tabular}}
  \caption{HumanEval results across sampling budgets ($n$), comparing Embedding-Based Agreement (EBA), Universal Self-Consistency (USC), Self-Certainty (SCe), and random selection for Llama 8B and Qwen 8B.  Results show accuracy scores multiplied by 100 with corresponding standard deviation as  a function of the number of sampled generations ($n$).}
  \label{tab:humaneval}
\end{table*}

\begin{table*}[ht]
  \centering
  \small
   \resizebox{\textwidth}{!}{
  \begin{tabular}{llcccccccc}
    \toprule
    \textbf{Model} & \textbf{Method} & $n=1$ & $n=4$ & $n=8$ & $n=16$ & $n=32$ & $n=64$ & $n=128$ & $n=256$ \\
    \midrule
    \textbf{Llama 8B} & EBA &52.77 $\pm$ \scriptsize{1.56} & 57.60 $\pm$ \scriptsize{1.37} & 60.03 $\pm$ \scriptsize{1.23} & 61.92 $\pm$ \scriptsize{1.11} & 63.13 $\pm$ \scriptsize{1.22} & 63.84 $\pm$ \scriptsize{1.25} & 64.39 $\pm$ \scriptsize{1.20} & 63.85 $\pm$ \scriptsize{1.05} \\
     & SCe & 52.77 $\pm$ \scriptsize{1.56}  & 54.57 $\pm$ \scriptsize{0.52} & 54.91 $\pm$ \scriptsize{0.41} & 54.95 $\pm$ \scriptsize{0.36} & 54.59 $\pm$ \scriptsize{0.43} & 54.00 $\pm$ \scriptsize{0.36} & 53.61 $\pm$ \scriptsize{0.42} & 51.65 $\pm$ \scriptsize{0.16} \\
     & USC & 52.77 $\pm$ \scriptsize{1.56}  & 57.50 $\pm$ \scriptsize{1.33} & 59.66 $\pm$ \scriptsize{1.35} & 60.02 $\pm$ \scriptsize{1.44} & -- & -- & -- & -- \\
     & Random & 52.77  & 52.77 & 52.77 & 52.77 & 52.77 & 52.77 & 52.77 & 52.77 \\
    \midrule
    \textbf{Qwen 8B} & EBA &  84.46 $\pm$ \scriptsize{1.09} & 86.72 $\pm$ \scriptsize{1.15} & 87.78 $\pm$ \scriptsize{0.90} & 88.65 $\pm$ \scriptsize{0.73} & 89.18 $\pm$ \scriptsize{0.60} & 89.67 $\pm$ \scriptsize{0.76} & 89.70 $\pm$ \scriptsize{0.63} & 89.85 $\pm$ \scriptsize{0.51} \\
     & SCe & 84.46 $\pm$ \scriptsize{1.09} & 86.74 $\pm$ \scriptsize{0.13} & 87.05 $\pm$ \scriptsize{0.20} & 86.99 $\pm$ \scriptsize{0.21} & 87.24 $\pm$ \scriptsize{0.16} & 87.10 $\pm$ \scriptsize{0.10} & 86.88 $\pm$ \scriptsize{0.11} & 86.93 $\pm$ \scriptsize{0.15} \\
     & USC & 84.46 $\pm$ \scriptsize{1.09} & 88.28 $\pm$ \scriptsize{0.94} & 88.29 $\pm$ \scriptsize{0.81} & -- & -- & -- & -- & -- \\
     & Random & 84.46 & 84.46 & 84.46 & 84.46 & 84.46 & 84.46 & 84.46 & 84.46 \\
    \bottomrule
  \end{tabular}}
  \caption{MATH500 results across sampling budgets ($n$), comparing Embedding-Based Agreement (EBA), Universal Self-Consistency (USC), Self-Certainty (SCe), and random selection for Llama 8B and Qwen 8B. Results show accuracy scores multiplied by 100 with corresponding standard deviation as  a function of the number of sampled generations ($n$).}
  \label{tab:math500}
\end{table*}

\begin{table*}[ht]
  \centering
  \small
   \resizebox{\textwidth}{!}{
  \begin{tabular}{llcccccccc}
    \toprule
    \textbf{Model} & \textbf{Method} & $n=1$ & $n=4$ & $n=8$ & $n=16$ & $n=32$ & $n=64$ & $n=128$ & $n=256$ \\
    \midrule
    \textbf{Llama 8B} & EBA & 29.21 $\pm$ \scriptsize{0.41} & 29.47 $\pm$ \scriptsize{0.10} & 29.58 $\pm$ \scriptsize{0.11} & 29.74 $\pm$ \scriptsize{0.12} & 29.81 $\pm$ \scriptsize{0.11} & 29.89 $\pm$ \scriptsize{0.10} & 29.96 $\pm$ \scriptsize{0.11} & 30.02 $\pm$ \scriptsize{0.10} \\

     & SCe & 29.20 $\pm$ \scriptsize{0.41} & 29.50 $\pm$ \scriptsize{0.04} & 29.60 $\pm$ \scriptsize{0.03} & 29.80 $\pm$ \scriptsize{0.03} & 29.80 $\pm$ \scriptsize{0.04} & 29.90 $\pm$ \scriptsize{0.02} & 29.80 $\pm$ \scriptsize{0.03} & 29.80 $\pm$ \scriptsize{0.02} \\
     & USC & 29.20 $\pm$ \scriptsize{0.41} & 28.80 $\pm$ \scriptsize{0.14} & 28.40 $\pm$ \scriptsize{0.11} & 28.20 $\pm$ \scriptsize{0.12} & 28.20 $\pm$ \scriptsize{0.13} & 28.20 $\pm$ \scriptsize{0.13} & 28.30 $\pm$ \scriptsize{0.11} & 28.60 $\pm$ \scriptsize{0.11} \\
     & Random & 29.20  & 29.20  & 29.20 & 29.20  & 29.20  & 29.20 & 29.20 & 29.20 \\
    \midrule
    \textbf{Qwen 8B} & EBA &  27.79 $\pm$ \scriptsize{0.10} & 27.87 $\pm$ \scriptsize{0.08} & 27.94 $\pm$ \scriptsize{0.09} & 28.03 $\pm$ \scriptsize{0.07} & 28.07 $\pm$ \scriptsize{0.07} & 28.11 $\pm$ \scriptsize{0.06} & 28.12 $\pm$ \scriptsize{0.07} & 28.19 $\pm$ \scriptsize{0.08} \\
     & SCe & 27.80 $\pm$ \scriptsize{0.10} & 27.80 $\pm$ \scriptsize{0.01} & 27.80 $\pm$ \scriptsize{0.02} & 27.80 $\pm$ \scriptsize{0.01} & 27.80 $\pm$ \scriptsize{0.01} & 27.70 $\pm$ \scriptsize{0.01} & 27.70 $\pm$ \scriptsize{0.01} & 27.70 $\pm$ \scriptsize{0.02} \\
     & USC & 27.80 $\pm$ \scriptsize{0.10} & 27.60 $\pm$ \scriptsize{0.09} & 27.50 $\pm$ \scriptsize{0.09} & 27.60 $\pm$ \scriptsize{0.09} & 27.60 $\pm$ \scriptsize{0.09} & 27.70 $\pm$ \scriptsize{0.10} & 27.60 $\pm$ \scriptsize{0.07} & 27.70 $\pm$ \scriptsize{0.09} \\
     & Random & 27.80 & 27.80 &27.80 & 27.80 & 27.80 & 27.80 &27.80 & 27.80 \\
    \bottomrule
  \end{tabular}}
   \caption{CNN/DM results across sampling budgets ($n$), comparing Embedding-Based Agreement (EBA), Universal Self-Consistency (USC), Self-Certainty (SCe), and random selection for Llama 8B and Qwen 8B.  Results show ROUGE-1 scores multiplied by 100 with corresponding standard deviation as  a function of the number of sampled generations ($n$).}
  \label{tab:cnn}
\end{table*}

\begin{table*}[ht]
  \centering
  \small
    \resizebox{\textwidth}{!}{
  \begin{tabular}{llcccccccc}
    \toprule
  \textbf{Model} & \textbf{Embedding} & $n=1$ & $n=4$ & $n=8$ & $n=16$ & $n=32$ & $n=64$ & $n=128$ & $n=256$ \\
    \midrule
  \multirow{3}{*}{\textbf{Llama 8B}} & Qwen Emb. & 65.98 $\pm$ \scriptsize{2.38} & 67.89 $\pm$ \scriptsize{2.23} & 68.72 $\pm$ \scriptsize{2.12} & 69.00 $\pm$ \scriptsize{1.97} & 69.82 $\pm$ \scriptsize{1.54} & 70.01 $\pm$ \scriptsize{1.96} & 70.11 $\pm$ \scriptsize{1.52} & 70.06 $\pm$ \scriptsize{1.58} \\
   & Gemma Emb. & 65.98 $\pm$ \scriptsize{2.38} & 68.29 $\pm$ \scriptsize{2.21} & 69.34 $\pm$ \scriptsize{2.15} & 69.83 $\pm$ \scriptsize{1.64} & 69.83 $\pm$ \scriptsize{2.07} & 70.34 $\pm$ \scriptsize{1.84} & 70.60 $\pm$ \scriptsize{1.24} & 70.30 $\pm$ \scriptsize{1.50} \\
   & Native Emb. & 65.98 $\pm$ \scriptsize{2.38} & 67.51 $\pm$ \scriptsize{2.34} & 68.07 $\pm$ \scriptsize{2.06} & 68.63 $\pm$ \scriptsize{1.92} & 68.70 $\pm$ \scriptsize{1.56} & 67.87 $\pm$ \scriptsize{1.79} & 67.88 $\pm$ \scriptsize{1.50} & 67.70 $\pm$ \scriptsize{1.56} \\
  \midrule
  \multirow{3}{*}{\textbf{Llama 70B}} & Qwen Emb. & 79.89 $\pm$ \scriptsize{1.88} & 80.39 $\pm$ \scriptsize{1.63} & 81.18 $\pm$ \scriptsize{1.76} & 81.87 $\pm$ \scriptsize{1.50} & 81.88 $\pm$ \scriptsize{1.40} & 82.06 $\pm$ \scriptsize{1.44} & 81.67 $\pm$ \scriptsize{1.32} & 80.61 $\pm$ \scriptsize{1.34} \\
   & Gemma Emb. & 79.89 $\pm$ \scriptsize{1.88} & 80.77 $\pm$ \scriptsize{1.52} & 81.30 $\pm$ \scriptsize{1.81} & 81.85 $\pm$ \scriptsize{1.49} & 82.24 $\pm$ \scriptsize{1.47} & 81.99 $\pm$ \scriptsize{1.30} & 82.28 $\pm$ \scriptsize{1.19} & 82.44 $\pm$ \scriptsize{1.08} \\
   & Native Emb. & 79.89 $\pm$ \scriptsize{1.88} & 80.41 $\pm$ \scriptsize{1.88} & 80.93 $\pm$ \scriptsize{1.62} & 81.49 $\pm$ \scriptsize{1.50} & 81.96 $\pm$ \scriptsize{1.64} & 82.88 $\pm$ \scriptsize{1.59} & 82.76 $\pm$ \scriptsize{1.29} & 83.02 $\pm$ \scriptsize{1.21} \\
  \midrule \midrule
  \multirow{3}{*}{\textbf{Qwen 8B}} & Qwen Emb. & 83.45 $\pm$ \scriptsize{2.10} & 86.07 $\pm$ \scriptsize{1.85} & 86.17 $\pm$ \scriptsize{1.46} & 87.32 $\pm$ \scriptsize{1.17} & 87.29 $\pm$ \scriptsize{1.26} & 88.00 $\pm$ \scriptsize{1.38} & 87.95 $\pm$ \scriptsize{1.21} & 88.11 $\pm$ \scriptsize{1.31} \\
   & Gemma Emb. & 83.45 $\pm$ \scriptsize{2.10} & 85.72 $\pm$ \scriptsize{1.69} & 85.88 $\pm$ \scriptsize{1.39} & 87.17 $\pm$ \scriptsize{1.23} & 87.09 $\pm$ \scriptsize{1.36} & 86.94 $\pm$ \scriptsize{1.05} & 86.84 $\pm$ \scriptsize{1.03} & 86.76 $\pm$ \scriptsize{0.90} \\
   & Native Emb. & 83.45 $\pm$ \scriptsize{2.10} & 85.80 $\pm$ \scriptsize{1.81} & 86.61 $\pm$ \scriptsize{1.45} & 86.61 $\pm$ \scriptsize{1.19} & 86.74 $\pm$ \scriptsize{1.20} & 86.71 $\pm$ \scriptsize{1.13} & 86.63 $\pm$ \scriptsize{1.19} & 87.02 $\pm$ \scriptsize{1.14} \\
  \midrule
  \multirow{3}{*}{\textbf{Qwen 32B}} & Qwen Emb. & 87.16 $\pm$ \scriptsize{1.87} & 89.01 $\pm$ \scriptsize{1.52} & 88.94 $\pm$ \scriptsize{1.30} & 89.70 $\pm$ \scriptsize{1.39} & 89.85 $\pm$ \scriptsize{1.08} & 90.07 $\pm$ \scriptsize{0.98} & 90.23 $\pm$ \scriptsize{0.99} & 90.76 $\pm$ \scriptsize{0.99} \\
   & Gemma Emb. & 87.16 $\pm$ \scriptsize{1.87} & 88.83 $\pm$ \scriptsize{1.26} & 89.26 $\pm$ \scriptsize{1.28} & 89.71 $\pm$ \scriptsize{1.17} & 89.54 $\pm$ \scriptsize{1.05} & 89.55 $\pm$ \scriptsize{1.08} & 89.52 $\pm$ \scriptsize{0.88} & 89.73 $\pm$ \scriptsize{0.82} \\
   & Native Emb. & 87.16 $\pm$ \scriptsize{1.87} & 88.52 $\pm$ \scriptsize{1.27} & 89.00 $\pm$ \scriptsize{1.41} & 89.38 $\pm$ \scriptsize{1.04} & 89.66 $\pm$ \scriptsize{1.28} & 89.13 $\pm$ \scriptsize{1.03} & 89.22 $\pm$ \scriptsize{1.26} & 89.66 $\pm$ \scriptsize{1.17} \\
    \bottomrule
  \end{tabular}}
   \caption{EBA performance on HumanEval across model scales (Llama 8B, Llama 70B, Qwen 8B, Qwen 32B) and embedding spaces, including Qwen embeddings, Gemma embeddings, and native hidden representations. Results show accuracy scores multiplied by 100 with corresponding standard deviation as  a function of the number of sampled generations ($n$).}
  \label{tab:eba_humaneval}
\end{table*}

\begin{table*}[ht]
  \centering
  \small
  \resizebox{\textwidth}{!}{
  \begin{tabular}{llcccccccc}
    \toprule
    \textbf{Model} & \textbf{Embedding} & $n=1$ & $n=4$ & $n=8$ & $n=16$ & $n=32$ & $n=64$ & $n=128$ & $n=256$ \\
    \midrule
  \multirow{3}{*}{\textbf{Llama 8B}} & Qwen Emb. & 52.77 $\pm$ \scriptsize{1.56} & 57.60 $\pm$ \scriptsize{1.37} & 60.03 $\pm$ \scriptsize{1.23} & 61.92 $\pm$ \scriptsize{1.11} & 63.13 $\pm$ \scriptsize{1.22} & 63.84 $\pm$ \scriptsize{1.25} & 64.39 $\pm$ \scriptsize{1.20} & 63.85 $\pm$ \scriptsize{1.05} \\
   & Gemma Emb. & 52.77 $\pm$ \scriptsize{1.56} & 56.71 $\pm$ \scriptsize{1.30} & 59.39 $\pm$ \scriptsize{1.16} & 61.57 $\pm$ \scriptsize{1.60} & 62.57 $\pm$ \scriptsize{1.30} & 63.58 $\pm$ \scriptsize{1.42} & 64.17 $\pm$ \scriptsize{1.17} & 65.14 $\pm$ \scriptsize{1.06} \\
   & Native Emb. & 52.77 $\pm$ \scriptsize{1.56} & 56.19 $\pm$ \scriptsize{1.20} & 58.68 $\pm$ \scriptsize{1.52} & 60.51 $\pm$ \scriptsize{1.17} & 61.57 $\pm$ \scriptsize{1.26} & 61.96 $\pm$ \scriptsize{1.04} & 62.04 $\pm$ \scriptsize{1.13} & 62.06 $\pm$ \scriptsize{0.75} \\
  \midrule
  \multirow{3}{*}{\textbf{Llama 70B}} & Qwen Emb. & 70.04 $\pm$ \scriptsize{1.44} & 73.15 $\pm$ \scriptsize{1.29} & 75.34 $\pm$ \scriptsize{1.16} & 76.92 $\pm$ \scriptsize{1.05} & 77.64 $\pm$ \scriptsize{1.05} & 78.29 $\pm$ \scriptsize{0.98} & 78.53 $\pm$ \scriptsize{0.84} & 78.83 $\pm$ \scriptsize{0.85} \\
   & Gemma Emb. & 70.04 $\pm$ \scriptsize{1.44} & 72.81 $\pm$ \scriptsize{1.31} & 74.68 $\pm$ \scriptsize{1.34} & 75.68 $\pm$ \scriptsize{1.06} & 76.20 $\pm$ \scriptsize{1.11} & 76.74 $\pm$ \scriptsize{1.04} & 76.96 $\pm$ \scriptsize{0.92} & 76.60 $\pm$ \scriptsize{1.04} \\
   & Native Emb. & 70.04 $\pm$ \scriptsize{1.44} & 73.46 $\pm$ \scriptsize{1.33} & 75.69 $\pm$ \scriptsize{0.96} & 77.00 $\pm$ \scriptsize{1.03} & 77.71 $\pm$ \scriptsize{0.96} & 78.30 $\pm$ \scriptsize{1.03} & 78.82 $\pm$ \scriptsize{0.74} & 78.77 $\pm$ \scriptsize{0.71} \\
  \midrule\midrule
  \multirow{3}{*}{\textbf{Qwen 8B}} & Qwen Emb. & 84.46 $\pm$ \scriptsize{1.09} & 86.72 $\pm$ \scriptsize{1.15} & 87.78 $\pm$ \scriptsize{0.90} & 88.65 $\pm$ \scriptsize{0.73} & 89.18 $\pm$ \scriptsize{0.60} & 89.67 $\pm$ \scriptsize{0.76} & 89.70 $\pm$ \scriptsize{0.63} & 89.85 $\pm$ \scriptsize{0.51} \\
   & Gemma Emb. & 84.46 $\pm$ \scriptsize{1.09} & 85.77 $\pm$ \scriptsize{0.91} & 86.49 $\pm$ \scriptsize{0.96} & 87.02 $\pm$ \scriptsize{0.82} & 87.55 $\pm$ \scriptsize{0.69} & 87.88 $\pm$ \scriptsize{0.78} & 87.97 $\pm$ \scriptsize{0.73} & 88.43 $\pm$ \scriptsize{0.68} \\
   & Native Emb. & 84.46 $\pm$ \scriptsize{1.09} & 85.60 $\pm$ \scriptsize{0.96} & 86.30 $\pm$ \scriptsize{1.02} & 86.94 $\pm$ \scriptsize{0.75} & 87.19 $\pm$ \scriptsize{0.82} & 87.48 $\pm$ \scriptsize{0.74} & 86.96 $\pm$ \scriptsize{0.77} & 86.23 $\pm$ \scriptsize{0.64} \\
  \midrule
  \multirow{3}{*}{\textbf{Qwen 32B}} & Qwen Emb. & 86.57 $\pm$ \scriptsize{1.08} & 89.03 $\pm$ \scriptsize{0.97} & 90.03 $\pm$ \scriptsize{0.82} & 90.68 $\pm$ \scriptsize{0.55} & 91.21 $\pm$ \scriptsize{0.62} & 91.25 $\pm$ \scriptsize{0.50} & 91.55 $\pm$ \scriptsize{0.47} & 91.38 $\pm$ \scriptsize{0.49} \\
   & Gemma Emb. & 86.57 $\pm$ \scriptsize{1.08} & 88.14 $\pm$ \scriptsize{0.82} & 88.78 $\pm$ \scriptsize{0.81} & 89.35 $\pm$ \scriptsize{0.90} & 89.92 $\pm$ \scriptsize{0.74} & 90.06 $\pm$ \scriptsize{0.71} & 90.18 $\pm$ \scriptsize{0.67} & 90.27 $\pm$ \scriptsize{0.70} \\
   & Native Emb. & 86.57 $\pm$ \scriptsize{1.08} & 87.92 $\pm$ \scriptsize{1.01} & 88.94 $\pm$ \scriptsize{0.87} & 89.40 $\pm$ \scriptsize{1.00} & 89.87 $\pm$ \scriptsize{0.73} & 89.91 $\pm$ \scriptsize{0.70} & 90.02 $\pm$ \scriptsize{0.64} & 89.90 $\pm$ \scriptsize{0.54} \\
    \bottomrule
  \end{tabular}}
   \caption{EBA performance on MATH500 across model scales (Llama 8B, Llama 70B, Qwen 8B, Qwen 32B) and embedding spaces, including Qwen embeddings, Gemma embeddings, and native hidden representations. Results show accuracy scores multiplied by 100 with corresponding standard deviation as  a function of the number of sampled generations ($n$).}
  \label{tab:eba_math500}
\end{table*}

\begin{table*}[ht]
  \centering
  \small
  \resizebox{\textwidth}{!}{

\begin{tabular}{llcccccccc}
    \toprule
  \textbf{Model} & \textbf{Embedding} & $n=1$ & $n=4$ & $n=8$ & $n=16$ & $n=32$ & $n=64$ & $n=128$ & $n=256$ \\
    \midrule
  \multirow{3}{*}{\textbf{Llama 8B}} & Qwen Emb. & 29.21 $\pm$ \scriptsize{0.41} & 29.47 $\pm$ \scriptsize{0.10} & 29.58 $\pm$ \scriptsize{0.11} & 29.74 $\pm$ \scriptsize{0.12} & 29.81 $\pm$ \scriptsize{0.11} & 29.89 $\pm$ \scriptsize{0.10} & 29.96 $\pm$ \scriptsize{0.11} & 30.02 $\pm$ \scriptsize{0.10} \\
   & Gemma Emb. & 29.21 $\pm$ \scriptsize{0.41} & 29.33 $\pm$ \scriptsize{0.12} & 29.46 $\pm$ \scriptsize{0.11} & 29.57 $\pm$ \scriptsize{0.09} & 29.62 $\pm$ \scriptsize{0.09} & 29.68 $\pm$ \scriptsize{0.09} & 29.67 $\pm$ \scriptsize{0.09} & 29.67 $\pm$ \scriptsize{0.08} \\
   & Native Emb. & 29.21 $\pm$ \scriptsize{0.41} & 29.25 $\pm$ \scriptsize{0.09} & 29.38 $\pm$ \scriptsize{0.12} & 29.50 $\pm$ \scriptsize{0.12} & 29.58 $\pm$ \scriptsize{0.11} & 29.61 $\pm$ \scriptsize{0.08} & 29.67 $\pm$ \scriptsize{0.08} & 29.75 $\pm$ \scriptsize{0.08} \\
  \midrule
  \multirow{3}{*}{\textbf{Llama 70B}} & Qwen Emb. & 29.48 $\pm$ \scriptsize{0.44} & 29.72 $\pm$ \scriptsize{0.11} & 29.87 $\pm$ \scriptsize{0.12} & 29.95 $\pm$ \scriptsize{0.10} & 30.02 $\pm$ \scriptsize{0.10} & 30.08 $\pm$ \scriptsize{0.10} & 30.17 $\pm$ \scriptsize{0.08} & 30.25 $\pm$ \scriptsize{0.09} \\
   & Gemma Emb. & 29.48 $\pm$ \scriptsize{0.44} & 29.63 $\pm$ \scriptsize{0.11} & 29.79 $\pm$ \scriptsize{0.09} & 29.89 $\pm$ \scriptsize{0.11} & 29.96 $\pm$ \scriptsize{0.10} & 29.99 $\pm$ \scriptsize{0.09} & 30.01 $\pm$ \scriptsize{0.09} & 30.00 $\pm$ \scriptsize{0.08} \\
   & Native Emb. & 29.48 $\pm$ \scriptsize{0.44} & 29.52 $\pm$ \scriptsize{0.14} & 29.68 $\pm$ \scriptsize{0.10} & 29.78 $\pm$ \scriptsize{0.11} & 29.86 $\pm$ \scriptsize{0.10} & 29.92 $\pm$ \scriptsize{0.10} & 29.99 $\pm$ \scriptsize{0.07} & 30.02 $\pm$ \scriptsize{0.07} \\
  \midrule\midrule
  \multirow{3}{*}{\textbf{Qwen 8B}} & Qwen Emb. & 27.79 $\pm$ \scriptsize{0.10} & 27.87 $\pm$ \scriptsize{0.08} & 27.94 $\pm$ \scriptsize{0.09} & 28.03 $\pm$ \scriptsize{0.07} & 28.07 $\pm$ \scriptsize{0.07} & 28.11 $\pm$ \scriptsize{0.06} & 28.12 $\pm$ \scriptsize{0.07} & 28.19 $\pm$ \scriptsize{0.08} \\
   & Gemma Emb. & 27.79 $\pm$ \scriptsize{0.10} & 27.87 $\pm$ \scriptsize{0.08} & 27.92 $\pm$ \scriptsize{0.09} & 27.97 $\pm$ \scriptsize{0.07} & 27.97 $\pm$ \scriptsize{0.08} & 28.00 $\pm$ \scriptsize{0.07} & 28.03 $\pm$ \scriptsize{0.06} & 28.08 $\pm$ \scriptsize{0.07} \\
   & Native Emb. & 27.79 $\pm$ \scriptsize{0.10} & 27.81 $\pm$ \scriptsize{0.09} & 27.87 $\pm$ \scriptsize{0.09} & 27.92 $\pm$ \scriptsize{0.07} & 27.94 $\pm$ \scriptsize{0.07} & 27.93 $\pm$ \scriptsize{0.09} & 27.95 $\pm$ \scriptsize{0.06} & 27.95 $\pm$ \scriptsize{0.09} \\
  \midrule
  \multirow{3}{*}{\textbf{Qwen 32B}} & Qwen Emb. & 25.49 $\pm$ \scriptsize{0.12} & 25.49 $\pm$ \scriptsize{0.10} & 25.51 $\pm$ \scriptsize{0.10} & 25.48 $\pm$ \scriptsize{0.11} & 25.49 $\pm$ \scriptsize{0.08} & 25.47 $\pm$ \scriptsize{0.12} & 25.48 $\pm$ \scriptsize{0.11} & 25.45 $\pm$ \scriptsize{0.10} \\
   & Gemma Emb. & 25.49 $\pm$ \scriptsize{0.12} & 25.48 $\pm$ \scriptsize{0.10} & 25.48 $\pm$ \scriptsize{0.11} & 25.49 $\pm$ \scriptsize{0.08} & 25.49 $\pm$ \scriptsize{0.11} & 25.49 $\pm$ \scriptsize{0.09} & 25.47 $\pm$ \scriptsize{0.09} & 25.47 $\pm$ \scriptsize{0.11} \\
   & Native Emb. & 25.49 $\pm$ \scriptsize{0.12} & 25.49 $\pm$ \scriptsize{0.09} & 25.51 $\pm$ \scriptsize{0.08} & 25.51 $\pm$ \scriptsize{0.11} & 25.48 $\pm$ \scriptsize{0.08} & 25.50 $\pm$ \scriptsize{0.10} & 25.49 $\pm$ \scriptsize{0.08} & 25.45 $\pm$ \scriptsize{0.09} \\
    \bottomrule
  \end{tabular}}
\caption{EBA performance on CNN/DM across model scales (Llama 8B, Llama 70B, Qwen 8B, Qwen 32B) and embedding spaces, including Qwen embeddings, Gemma embeddings, and native hidden representations. Results show ROUGE-1 scores multiplied by 100 with corresponding standard deviation as a function of the number of sampled generations ($n$).}
\label{tab:eba_cnn_dm}
\end{table*}

\begin{table*}[ht]
  \centering
  \small
  \resizebox{\textwidth}{!}{
  \begin{tabular}{llcccccccc}
    \toprule
    \textbf{Model} & \textbf{Method} & $n=1$ & $n=4$ & $n=8$ & $n=16$ & $n=32$ & $n=64$ & $n=128$ & $n=256$ \\
    \midrule
    
    \multirow{3}{*}{\textbf{Llama 8B}}
    & EBA (Full Ans)
    & 52.77 $\pm$ \scriptsize{1.56}
    & 57.60 $\pm$ \scriptsize{1.37}
    & 60.03 $\pm$ \scriptsize{1.23}
    & 61.92 $\pm$ \scriptsize{1.11}
    & 63.13 $\pm$ \scriptsize{1.22}
    & 63.84 $\pm$ \scriptsize{1.25}
    & 64.39 $\pm$ \scriptsize{1.20}
    & 63.85 $\pm$ \scriptsize{1.05} \\
    
    & EBA (Extracted Ans)
    & 52.77 $\pm$ \scriptsize{1.56}
    & 60.06 $\pm$ \scriptsize{1.43}
    & 63.66 $\pm$ \scriptsize{1.24}
    & 66.36 $\pm$ \scriptsize{1.12}
    & 67.53 $\pm$ \scriptsize{1.02}
    & 68.41 $\pm$ \scriptsize{1.04}
    & 68.90 $\pm$ \scriptsize{1.01}
    & 69.20 $\pm$ \scriptsize{0.80} \\
    
    & SC
    & 52.77 $\pm$ \scriptsize{1.56}
    & 58.22 $\pm$ \scriptsize{0.98}
    & 63.85 $\pm$ \scriptsize{1.29}
    & 67.32 $\pm$ \scriptsize{1.14}
    & 69.36 $\pm$ \scriptsize{1.02}
    & 70.57 $\pm$ \scriptsize{0.98}
    & 71.23 $\pm$ \scriptsize{0.78}
    & 71.90 $\pm$ \scriptsize{0.70} \\
    
    \midrule
    
    \multirow{3}{*}{\textbf{Qwen 8B}}
    & EBA (Full Ans)
    & 84.46 $\pm$ \scriptsize{1.09}
    & 86.72 $\pm$ \scriptsize{1.15}
    & 87.78 $\pm$ \scriptsize{0.90}
    & 88.65 $\pm$ \scriptsize{0.73}
    & 89.18 $\pm$ \scriptsize{0.60}
    & 89.67 $\pm$ \scriptsize{0.76}
    & 89.70 $\pm$ \scriptsize{0.63}
    & 89.85 $\pm$ \scriptsize{0.51} \\
    
    & EBA (Extracted Ans)
    & 84.46 $\pm$ \scriptsize{1.09}
    & 87.73 $\pm$ \scriptsize{0.85}
    & 88.94 $\pm$ \scriptsize{0.88}
    & 89.58 $\pm$ \scriptsize{0.51}
    & 90.25 $\pm$ \scriptsize{0.64}
    & 90.42 $\pm$ \scriptsize{0.44}
    & 90.64 $\pm$ \scriptsize{0.46}
    & 90.51 $\pm$ \scriptsize{0.34} \\
    
    & SC
    & 84.46 $\pm$ \scriptsize{1.09}
    & 86.38 $\pm$ \scriptsize{0.94}
    & 88.84 $\pm$ \scriptsize{0.88}
    & 90.22 $\pm$ \scriptsize{0.78}
    & 90.62 $\pm$ \scriptsize{0.57}
    & 90.88 $\pm$ \scriptsize{0.48}
    & 91.16 $\pm$ \scriptsize{0.43}
    & 91.10 $\pm$ \scriptsize{0.34} \\
    
    \bottomrule
    \end{tabular}}
    \caption{Comparison of self-consistency (SC) and embedding-based agreement (EBA) using the complete generated response (EBA Full Ans) and only the extracted final answer (EBA Extracted Ans) on MATH500. Results are reported across different sample sizes ($n$) for Llama 8B and Qwen 8B models.  Results show accuracy scores multiplied by 100 with corresponding standard deviation as a function of the number of sampled generations $(n)$.}
\label{tab:matH_sc_8b}
\end{table*}

\begin{table*}[ht]
  \centering
  \small
  \resizebox{\textwidth}{!}{
  \begin{tabular}{llcccccccc}
    \toprule
    \textbf{Model} & \textbf{Method} & $n=1$ & $n=4$ & $n=8$ & $n=16$ & $n=32$ & $n=64$ & $n=128$ & $n=256$ \\
    \midrule
    
    \multirow{3}{*}{\textbf{Llama 70B}}
    & EBA (Full Ans)
    & 70.04 $\pm$ \scriptsize{1.44}
    & 73.15 $\pm$ \scriptsize{1.29}
    & 75.34 $\pm$ \scriptsize{1.16}
    & 76.92 $\pm$ \scriptsize{1.05}
    & 77.64 $\pm$ \scriptsize{1.05}
    & 78.29 $\pm$ \scriptsize{0.98}
    & 78.53 $\pm$ \scriptsize{0.84}
    & 78.83 $\pm$ \scriptsize{0.85} \\

    & EBA (Extracted Ans)
    & 70.04 $\pm$ \scriptsize{1.44}
    & 75.57 $\pm$ \scriptsize{1.24}
    & 77.74 $\pm$ \scriptsize{1.09}
    & 79.42 $\pm$ \scriptsize{0.83}
    & 80.51 $\pm$ \scriptsize{0.70}
    & 81.08 $\pm$ \scriptsize{0.66}
    & 81.30 $\pm$ \scriptsize{0.64}
    & 81.50 $\pm$ \scriptsize{0.46} \\

    & SC
    & 70.04 $\pm$ \scriptsize{1.44}
    & 74.86 $\pm$ \scriptsize{1.17}
    & 78.34 $\pm$ \scriptsize{0.87}
    & 79.99 $\pm$ \scriptsize{0.75}
    & 80.69 $\pm$ \scriptsize{0.77}
    & 81.10 $\pm$ \scriptsize{0.54}
    & 81.50 $\pm$ \scriptsize{0.51}
    & 81.72 $\pm$ \scriptsize{0.38} \\

    \midrule

    \multirow{3}{*}{\textbf{Qwen 32B}}
    & EBA (Full Ans)
    & 86.57 $\pm$ \scriptsize{1.08}
    & 89.03 $\pm$ \scriptsize{0.97}
    & 90.03 $\pm$ \scriptsize{0.82}
    & 90.68 $\pm$ \scriptsize{0.55}
    & 91.21 $\pm$ \scriptsize{0.62}
    & 91.25 $\pm$ \scriptsize{0.50}
    & 91.55 $\pm$ \scriptsize{0.47}
    & 91.38 $\pm$ \scriptsize{0.49} \\

    & EBA (Extracted Ans)
    & 86.57 $\pm$ \scriptsize{1.08}
    & 89.62 $\pm$ \scriptsize{0.90}
    & 90.68 $\pm$ \scriptsize{0.71}
    & 91.60 $\pm$ \scriptsize{0.41}
    & 92.41 $\pm$ \scriptsize{0.46}
    & 92.72 $\pm$ \scriptsize{0.31}
    & 92.99 $\pm$ \scriptsize{0.36}
    & 93.30 $\pm$ \scriptsize{0.27} \\

    & SC
    & 86.57 $\pm$ \scriptsize{1.08}
    & 88.86 $\pm$ \scriptsize{1.05}
    & 91.00 $\pm$ \scriptsize{0.75}
    & 92.03 $\pm$ \scriptsize{0.60}
    & 92.57 $\pm$ \scriptsize{0.45}
    & 92.90 $\pm$ \scriptsize{0.49}
    & 93.16 $\pm$ \scriptsize{0.31}
    & 93.28 $\pm$ \scriptsize{0.26} \\

    \bottomrule
    \end{tabular}}
    \caption{Comparison of self-consistency (SC) and embedding-based agreement (EBA) using the complete generated response (EBA Full Ans) and only the extracted final answer (EBA Extracted Ans) on MATH500. Results are reported across different sample sizes ($n$) for Llama 70B and Qwen 32B models.  Results show accuracy scores multiplied by 100 with corresponding standard deviation as a function of the number of sampled generations $(n)$.}
    \label{tab:matH_sc_70b}
\end{table*}

\begin{table*}[ht]
  \centering
  \small
  \resizebox{\textwidth}{!}{
  \begin{tabular}{llcccccccc}
    \toprule
    \textbf{Model} & \textbf{Method} & $n=1$ & $n=4$ & $n=8$ & $n=16$ & $n=32$ & $n=64$ & $n=128$ & $n=256$ \\
    \midrule
    
    \multirow{3}{*}{\textbf{Llama 8B}}
    & EBA 
    & 52.77 $\pm$ \scriptsize{1.56}
    & 57.60 $\pm$ \scriptsize{1.37}
    & 60.03 $\pm$ \scriptsize{1.23}
    & 61.92 $\pm$ \scriptsize{1.11}
    & 63.13 $\pm$ \scriptsize{1.22}
    & 63.84 $\pm$ \scriptsize{1.25}
    & 64.39 $\pm$ \scriptsize{1.20}
    & 63.85 $\pm$ \scriptsize{1.05} \\
    
    & Nearest &52.77&    57.50 & 60.05 & 61.35 & 62.98 & 63.62 & 64.12 & 63.87 \\
    
    & Farthest &52.77 & 46.99 & 42.78 & 40.10 & 37.59 & 35.71 & 32.60 & 29.98  \\
    & Random & 52.77  & 52.77 & 52.77 & 52.77 & 52.77 & 52.77 & 52.77 & 52.77 \\
    \midrule
    \multirow{3}{*}{\textbf{Qwen 8B}}
    & EBA 
    & 84.46 $\pm$ \scriptsize{1.09}
    & 86.72 $\pm$ \scriptsize{1.15}
    & 87.78 $\pm$ \scriptsize{0.90}
    & 88.65 $\pm$ \scriptsize{0.73}
    & 89.18 $\pm$ \scriptsize{0.60}
    & 89.67 $\pm$ \scriptsize{0.76}
    & 89.70 $\pm$ \scriptsize{0.63}
    & 89.85 $\pm$ \scriptsize{0.51} \\
     & Nearest &84.46  & 85.50 & 86.21 & 86.06 & 86.24 & 86.55 & 85.92 & 85.31 \\
    & Farthest &84.46  & 82.47 & 81.22 & 79.96 & 78.52 & 77.80 & 76.16 & 74.73   \\    
    & Random & 84.46 & 84.46 & 84.46 & 84.46 & 84.46 & 84.46 & 84.46 & 84.46 \\
    \bottomrule
    \end{tabular}}
    \caption{Comparison of generation selection strategies on MATH500 for Llama 8B and Qwen 8B across increasing numbers of sampled generations ($n$). EBA selects generations using agreement-based clustering, while \textit{Nearest} and \textit{Farthest} choose generations according to their distance from the global centroid in embedding space. \textit{Random} corresponds to the baseline performance at $n=1$. Results show accuracy scores multiplied by 100 with corresponding standard deviation (when possible) as a function of the number of sampled generations $(n)$.}
\label{tab:matH_centroid}
\end{table*}

\end{document}